\pdfoutput=1

\documentclass[11pt]{article}

\usepackage[]{acl}

\usepackage{times}
\usepackage{latexsym}

\usepackage[T1]{fontenc}

\usepackage[utf8]{inputenc}

\usepackage{microtype}

\usepackage{inconsolata}

\usepackage{multirow}
\usepackage{pdflscape}
\usepackage{amssymb}
\usepackage{amsmath}
\usepackage{amsfonts}
\usepackage{graphicx}
\usepackage{threeparttable}
\usepackage{booktabs}
\usepackage{xspace}
\usepackage{listings}
\definecolor{codegreen}{rgb}{0,0.6,0}
\definecolor{codegray}{rgb}{0.5,0.5,0.5}
\definecolor{codepurple}{rgb}{0.58,0,0.82}
\definecolor{backcolour}{rgb}{0.95,0.95,0.92}
\lstdefinestyle{mystyle}{
    backgroundcolor=\color{backcolour},   
    commentstyle=\color{codegreen},
    keywordstyle=\color{magenta},
    numberstyle=\tiny\color{codegray},
    stringstyle=\color{codepurple},
    basicstyle=\ttfamily\footnotesize,
    breakatwhitespace=false,         
    breaklines=true,                 
    captionpos=b,                    
    keepspaces=true,                 
    numbers=left,                    
    numbersep=5pt,                  
    showspaces=false,                
    showstringspaces=false,
    showtabs=false,                  
    tabsize=2
}
\newcommand{\ndjv}{\texttt{News D\'ej\`a Vu}\xspace}

\begin{document}

\title{\ndjv: Connecting Past and Present with Semantic Search} 
\author{Brevin Franklin, Emily Silcock, Abhishek Arora, Tom Bryan, and Melissa Dell$^{\ast}$ \\
Harvard University, Cambridge, MA, USA. Authors contributed equally. \\
\normalsize{$^\ast$Corresponding author:  melissadell@fas.harvard.edu.}
}
\maketitle

\begin{abstract}
Social scientists and the general public often analyze contemporary events by drawing parallels with the past, a process complicated by the vast, noisy, and unstructured nature of historical texts. For example, hundreds of millions of page scans from historical newspapers have been noisily transcribed. Traditional sparse methods for searching for relevant material in these vast corpora, \textit{e.g.,} with keywords, can be brittle given complex vocabularies and OCR noise. This study introduces \ndjv, a novel semantic search tool that leverages transformer large language models and a bi-encoder approach to identify historical news articles that are most similar to modern news queries. \ndjv first recognizes and masks entities, in order to focus on broader parallels rather than the specific named entities being discussed. Then, a contrastively trained, lightweight bi-encoder retrieves historical articles that are most similar semantically to a modern query, illustrating how phenomena that might seem unique to the present have varied historical precedents. Aimed at social scientists, the user-friendly \ndjv package is designed to be accessible for those who lack extensive familiarity with deep learning. It works with large text datasets, and we show how it can be deployed to a massive scale corpus of historical, open-source news articles. We curate some examples on \url{ newsdejavu.github.io}. While human expertise remains important for drawing deeper insights, \ndjv provides a powerful tool for exploring parallels in how people have perceived past and present.
\end{abstract}


\begin{quote}
    "Those who cannot remember the past
are condemned to repeat it."
-- George Santayana's
The Life of Reason
\end{quote}

\section{Introduction}

Social scientists, and the public more generally, often seek to place the present in perspective by reflecting upon parallels with the past. Finding these commonalities, however, can be a labor-intensive and challenging process. Vast troves of historical texts have been preserved, but they are often held in unstructured, uncataloged, massive-scale databases. For example, hundreds of millions of pages from historical newspapers have been digitized and are available online through both open-source and proprietary collections. Keyword searches are often used to extract relevant documents from these massive corpora. However, as language is complex and OCR noise is rampant, sparse methods can be extremely brittle.

\begin{figure*}
    \centering
    \includegraphics[width=\textwidth]{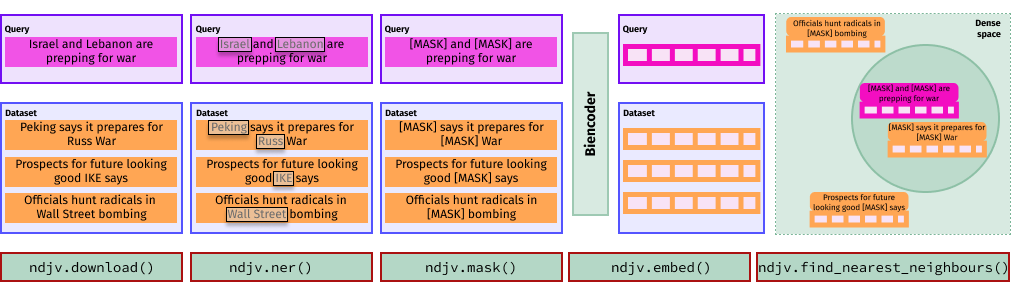}
    \caption{\ndjv architecture at inference time.}
    \label{fig:arch}
\end{figure*}

Transformer large language models offer a powerful tool for retrieving source material from the past that can contextualize the present. This study trains a novel semantic search model,  \ndjv, to query which historical news articles are most semantically similar to a modern news article query. Figure \ref{fig:arch} shows the model architecture at inference time. Named entities are first detected and masked out, using a named entity recognition model that we tuned for noisy, historical texts. This allows the model to focus on the generalities of the story, rather than specific names of people, locations, or organizations. Then, we use a contrastively trained bi-encoder model to retrieve the modern article's nearest neighbor(s) from a massive-scale database of historical texts. 

The \ndjv package allows social scientists to deploy \ndjv on their own queries. It has a CC-BY license and can be used with any appropriately formatted text dataset. It is designed to be user-friendly and intuitive to social scientists, who often lack knowledge of deep learning frameworks. This study provides code snippets showing how it can be used seamlessly with American Stories, a Hugging Face dataset containing over 430 million historical public domain newspaper article texts \cite{dell2023american}.  
Interested users can query content from a sampling of states in American Stories with modern articles using our HuggingFace demo.\footnote{ \url{https://huggingface.co/spaces/dell-research-harvard/newsdejavu}}
We also maintain a website (\url{newsdejavu.github.io}), where randomly selected modern news articles - as well as special editions on hand-curated topics of interest - are paired with their retrieved historical neighbors. 
We built the \ndjv package after many website readers requested that we create a tool that they could use to query their own texts of interest. 

\ndjv retrieves articles that use similar semantics. Of course, events or phenomena that are at their core very different may be described in a similar way by the historical and modern news media. This phenomenon is also likely to be of considerable interest to social scientists, but we do caveat that deeper historical knowledge is required to place parallels in appropriate context. 

\ndjv currently supports English. In the future, it could be relatively straightforward to create a multilingual model by starting with multilingual Sentence BERT weights and tuning on machine-translated allsides data or other data sources (\textit{e.g.}, \citet{chen2022semeval}). 

The rest of this study is organized as follows. Section \ref{sec:lit} discusses the relevant literature, and Section \ref{sec:architecture} describes the model architecture and training. Section \ref{sec:package} introduces the \ndjv package.

\section{Related Literature} \label{sec:lit}

There is a large literature on semantic similarity. Most large scale datasets in this space are constructed from web texts. The Massive Text Embedding Benchmark (MTEB) \cite{muennighoff2022mteb}, evaluates 8 embedding tasks on 58 datasets covering 112 languages, providing an overview of available datasets.

This study relates most closely to \citet{silcock2022noise}, which contrastively trains an S-BERT MPNet model \cite{reimers2019sentence, song2020mpnet} to map historical newswire articles from the same underlying article source to similar embeddings. 
We initialize \ndjv with their model weights.

More generally, this study follows from the literature on open domain retrieval \citep{karpukhin2020dense,thakur2021beir,wu2019scalable}. We also draw inspiration from a large literature showing the importance of contrastive training for semantic similarity applications, which we apply to train \ndjv. The anisotropic shape of the embedding space in pre-trained transformer models like BERT creates challenges for utilizing their latent features \cite{ethayarajh2019contextual}. In these models, less common words are dispersed towards the edges of the hypersphere, the sparsity of low frequency words violates convexity, and the distance between embeddings is correlated with lexical similarity. This leads to misalignment among texts with similar meanings and diminishes the effectiveness of averaging token embeddings to represent longer texts \cite{reimers2019sentence}. By applying contrastive training, anisotropy is mitigated \cite{wang2020understanding}, enhancing the quality of pooled sentence (or document) representations \cite{reimers2019sentence}.

\begin{figure*}
    \centering
    \includegraphics[width=0.85\textwidth]{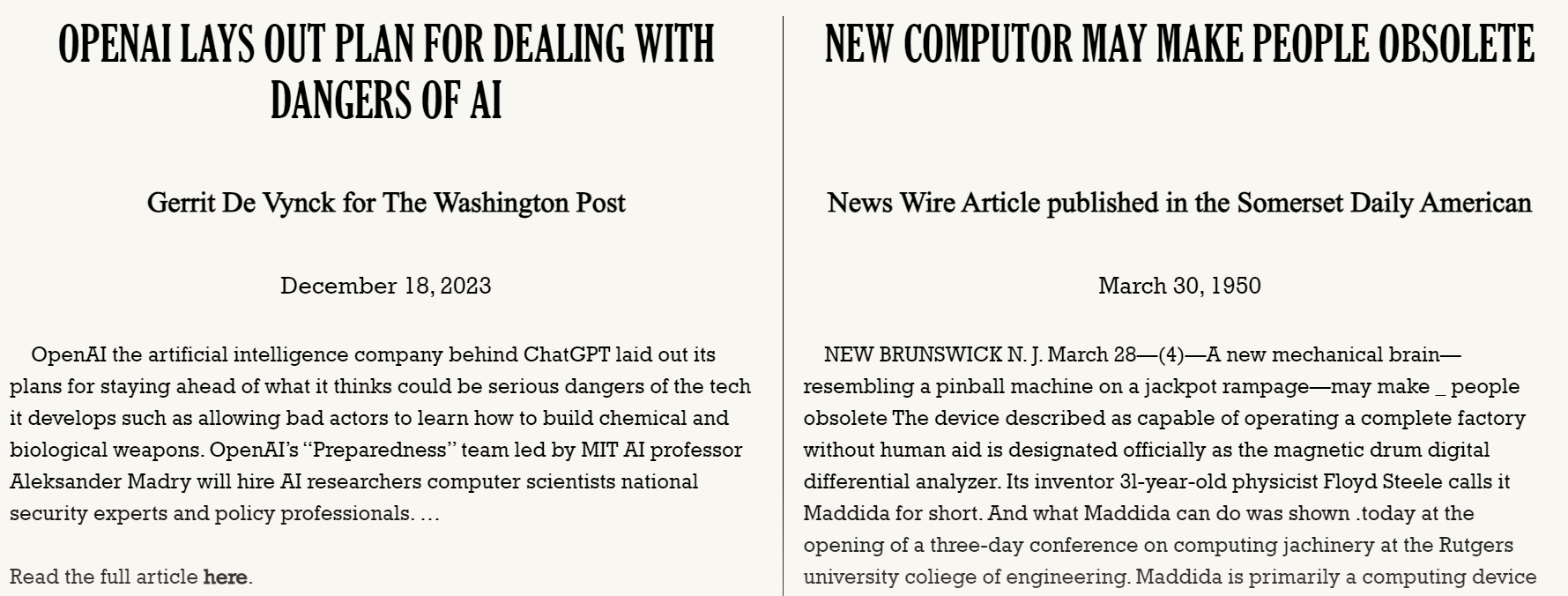}
    \includegraphics[width=0.85\textwidth]{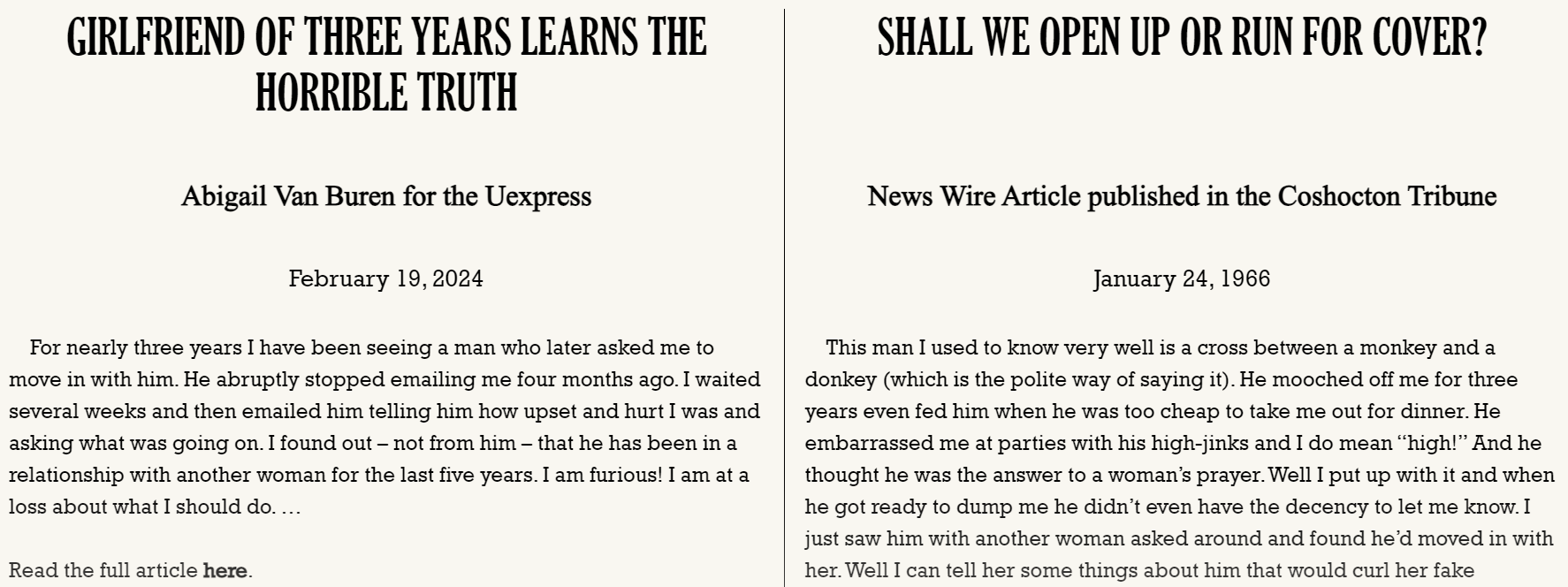}
    \caption{Examples of \ndjv retrieval. The left-hand side shows a modern news article and the right-hand side shows a retrieved historical article.}
    \label{fig:web_ex}
\end{figure*}

\section{Model Architecture and Training} \label{sec:architecture}

The \ndjv model architecture at inference time is shown in Figure \ref{fig:arch}. \ndjv first recognizes and masks spans of text containing named entities (people, locations, organizations, and other miscellaneous proper nouns), as our aim is to draw parallels between articles that describe different entities in different time periods. We then replace all detected entities by the [MASK] token. Query articles are used to retrieve their semantic nearest neighbor(s) in a corpus of interest, using the \ndjv contrastively trained bi-encoder. We use the IndexFlatIP index from FAISS \citep{johnson2019billion} to perform an exact K-nearest neighbor search. Our embedding vectors are L2 normalized which makes the Inner Product metric used in the index equivalent to Cosine Similarity.

Custom training was necessary to achieve accurate NER performance with robustness to OCR noise. Table \ref{tab:entity_distribution} describes the training data, which were drawn from randomly selected articles from off-copyright newspaper articles between 1922 and 1977. 
All data were double-labeled by two highly skilled undergraduate research assistants, and all discrepancies were resolved by hand. The supplemental materials contain the annotator instructions.
We use the training set to fine-tune a Roberta-Large model \citep{liu2019roberta}. We optimised hyperparameters using Hyperband \cite{li2018hyperband}. The best model was trained at a learning rate of 4.7e-05 with a batch size of 128 for 184 epochs. 

\begin{table}[ht]
\centering
\resizebox{\linewidth}{!}{\begin{tabular}{lccccc}
\hline
Split   & Person & Org & Loc & Misc & Articles \\ \hline
Train   & 1345   & 450          & 1191     & 1037 & 227            \\
Val     & 231    & 59           & 192      & 149  & 48             \\
Test    & 261    & 83           & 199      & 181  & 48             \\ \hline
\end{tabular}}
\caption{The first four columns provide the number of entities of different types in the training, validation, and test sets. The final column provides the total number of labeled articles.}
\label{tab:entity_distribution}
\end{table}

Table \ref{tab:entity_eval}  evaluates NER model performance. 
This model achieves an F1 of 90.4 in correctly identifying spans of text containing named entities without regards to the class, the relevant task since \ndjv replaces all entities with the [MASK] token. This outperforms a Roberta-Large model finetuned on CoNLL03 by a large margin. 

\begin{table}[ht]
\centering
\resizebox{\linewidth}{!}
{
\begin{tabular}{lccc}
\hline
Model & Precision & Recall & F1 \\ \hline
Custom NER & 87.9   & 93.1 & 90.4 \\
Roberta-Large finetuned on & 80.3   & 75.5 & 77.8 \\
\hspace{0.2cm} CoNLL03 \citep{conneau2019unsupervised} & \\

\hline
\end{tabular}}
\caption{Evaluation of NER models.}
\label{tab:entity_eval}
\end{table}

We would like to use \ndjv for unsupervised data exploration, to retrieve historical texts that social scientists and the general public will find thought-provoking.
In order to do this, we need an LLM that maps semantically similar articles to similar representations, and we found that off-the-shelf contrastively trained models - such as those described in the MTEB benchmark - did not perform satisfactorily.

Creating paired training data of modern and historical articles that do or do not have parallels would be challenging and costly.
Rather, we begin with the model from \citet{silcock2022noise}, which was contrastively trained on paired historical newswire articles, with the purpose of detecting noisy duplication, rather than semantic similarity. This is a useful starting point since it has already been exposed extensively to the idiosyncrasies of historical news texts, such as OCR errors and obsolete spellings. 

We further train on modern data that pairs news articles belonging to the same news story. These are drawn from \texttt{Allsides}, a news aggregator that collates the beginnings of articles on the same story from multiple news sites. Pairs of (the beginnings of) articles from these groupings, which typically consist of two or three texts, form positives. 
To create negative pairs, we used a larger pool of articles from \texttt{Allsides}, leveraging their pages of articles that are on the same topic, which are broader groupings than those on the same story. We embed this pool using the model from \citet{silcock2022noise}, which is a fintuned S-BERT  MPNET model \citep{reimers2019sentence,song2020mpnet}. Then for each article that appears in a positive pair, we find the closest article (highest cosine similarity) in the pool that a) is from the same news source and b) does not appear on the same topic or story page.\footnote{In some cases there was no article that met these conditions. In these cases, we took an article from another news source.} Training data statistics are given in Table \ref{tab:splits}.

\begin{table}[htbp]
    \centering
    \resizebox{\linewidth}{!}{
    \begin{threeparttable}
    \begin{tabular}{cccccccc}
        \toprule
        \multicolumn{2}{c}{\textbf{Training}} & \multicolumn{2}{c}{\textbf{Validation}} & \multicolumn{2}{c}{\textbf{Test}} & \multicolumn{2}{c}{\textbf{Total}}\\
         Pos. & Neg. & Pos. & Neg. & Pos. & Neg. & Pos. & Neg. \\
        \midrule
        12868 & 12913 & 2757 & 2766 & 2757  & 2766 & 18,382 & 18,445 \\
        \bottomrule
    \end{tabular}
    \end{threeparttable}
   }
    \caption{Training, validation, and test sizes for the paired data used to train the retriever.}
    \label{tab:splits}
\end{table}

We used Hyperband \cite{li2018hyperband} to find optimal hyperparameters, which led us to train for 9 epochs, with a batch size of 32 and a warm-up rate of 0.39. We use S-BERT’s online contrastive loss (Hadsell et al., 2006) implementation with a margin of 0.5. The model achieves a pairwise F1 of 92.4 on the test set, outperforming models that are not finetuned, such as SBERT MPNET \citep{reimers2019sentence} and \citet{silcock2022noise}, as shown in table \ref{tab:biencoder_eval}. 

\begin{table}[ht]
\centering
\resizebox{\linewidth}{!}
{
\begin{tabular}{lccc}
\hline
Model & Precision & Recall & F1 \\ \hline
Custom biencoder & 93.7 & 91.1   & 92.4 \\
\citet{reimers2019sentence} & 83.8 & 85.8   & 84.8 \\
\citet{silcock2022noise} & 60.7 & 69.5   & 64.8 \\

\hline
\end{tabular}}
\caption{Evaluation of biencoder models.}
\label{tab:biencoder_eval}
\end{table}

We do not have paired modern-historical article texts for evaluation, and it is not clear how one would create such data given the nature of the task. Rather, a skilled annotator gave 350 randomly selected modern articles and their nearest historical \ndjv neighbor a short description of their main topic (typically 2-3 words). In 85.7\% of cases, the modern query and its nearest historical neighbor had the same major topic. Even when the major topic is not the same, the pairs often showed other interesting parallels. 

Figure \ref{fig:web_ex} provides representative examples of \ndjv retrieval. Other examples can be seen at \url{newsdejavu.github.io}. Modern query articles are truncated due to copyright protection. Historical articles are drawn from off-copyright newspapers and reproduced in their entirety. Except for a few special editions about topics requested by our readers, these articles were selected at random, painting a representative picture of \ndjv. 

We have a demo where users can use their own texts to query a subset of American Stories \cite{dell2023american}, a massive scale Hugging Face dataset consisting of over 430 million historical newspaper texts.\footnote{\url{https://huggingface.co/spaces/dell-research-harvard/newsdejavu}} We also make embeddings for the American Stories collection available on Hugging Face.\footnote{\url{https://huggingface.co/datasets/dell-research-harvard/americanstories_masked_embeddings}}

\section{The \ndjv Package} \label{sec:package}

The \ndjv package is available on PyPI for easy install. 

\begin{lstlisting}[language=python]
pip install newsdejavu
\end{lstlisting}

The package consists of the following core functionalities: \texttt{download}, \texttt{ner}, \texttt{mask}, \texttt{embed}, and \texttt{find\_nearest\_neighbours}. The package focuses on inference, which we expect is how the vast majority of users would like to use \ndjv. 
For users who wish to fine-tune their own \ndjv model, we recommend using LinkTransformer \cite{arora2023linktransformer} or the Sentence BERT repository \cite{reimers2019sentence}, initializing with our pre-trained weights which are available on Hugging Face.

The \texttt{download} function downloads the dataset that users would like to work with. We have integrated support for American Stories \cite{dell2023american}. This command allows users to specify which states and year range they would like to download, or they can download the (very large) dataset in full. This step of the pipeline can also be skipped if the user already has a dataset that they would like to use. 

\begin{lstlisting}[language=python]
import newsdejavu as ndjv
corpus = ndjv.download('american stories:1900:Alabama')
\end{lstlisting}

The \texttt{ner} command runs NER over the corpus. 

\begin{lstlisting}[language=python]
ner_outputs = ndjv.ner(corpus,  'historical_newspaper_ner')
\end{lstlisting}

Next, these detected entities can be masked with a simple \texttt{mask} command and the texts embedded using \ndjv with the \texttt{embed} command. In addition to using the \ndjv model as the default, this command can also support using a local model path or downloading one from Hugging Face, for users who would like to use their own retrieval models in conjunction with the package. 

\begin{lstlisting}[language=python]
masked_corpus = ndjv.mask(ner_outputs)
corpus_embeddings = ndjv.embed(masked_corpus, 'same-story')
\end{lstlisting}

Users can similarly mask and embed their query article. Finally, \texttt{find\_nearest\_neighbours} retrieves the $k$ closest corpus articles to the query.

\begin{lstlisting}[language=python]
dist_list, nn_list = find_nearest_neighbours(query_embeddings, corpus_embeddings, k=1)
\end{lstlisting}

Users would typically like to use all these commands in sequence. The \texttt{mask\_and\_embed} command combines NER, masking, and embedding, and the \texttt{search\_nearest\_story} command combines NER, masking, embedding, and retrieval. 

\begin{lstlisting}[language=python]
corpus_embeddings = ndjv.mask_and_embed(ner_outputs)
nearest_articles = njdv.search_nearest_story(query_articles, 'historical_newspaper_ner', 'same-story', corpus_embed = corpus_embeddings)
\end{lstlisting}


We recommend that those who lack extensive familiarity with deep learning frameworks install it on a cloud compute service optimized for deep learning, such as Google Colab, in order to avoid the need to resolve dependencies. 
Tutorials on how to use \ndjv on Colab will be provided on the \ndjv Github Repository (\url{github.com/dell-research-harvard/newsdejavu/}) and on the website.

By making semantic search an accessible tool for social scientists to apply to historical document collections, we hope to make it easier for researchers to find content that contextualizes our understanding of the parallels between past and present.

\section*{Ethics Statement} \label{sec:ethics}
\ndjv is ethically sound. We do emphasize that it retrieves articles that use similar language, which may or may not reflect similarities in the underlying events or phenomena being described. Trained human judgement is required to draw deeper parallels between the past and present, and we hope \ndjv will be a useful tool for directing researchers and the public to content of interest. 

\section*{Acknowlegements} 
We are grateful for research assistance from Dennis Du, Jude Ha, Alice Liu, Shiloh Liu, Stephanie Lin, Andrew Lu, Prabhav Kamojjhala, and Ryan Xia.

\bibliography{cites}
\bibstyle{acl_natbib}

\clearpage

\appendix\section{NER annotator instructions and results}

During the NER annotation process, careful rules were developed to ensure congruence between use of labels. This appendix details those rules

\subsection{General rules}
\begin{itemize}
\item Label the biggest span than constitutes one entity, with the exception of locations. For example ``Martin Luther King High School'' would be one entity, not ``Martin Luther King'' or ``High School''. 

\item Label ``Vietnamese government'', not just ``Vietnamese''.

\item ``Adam Smith'' is one person and the ``Catholic Church'' is one organization, but ``Topeka, Kansas is two locations.

\item If the sub-entity is ambiguous without connection to the parent entity, we label it as one entity. For instance, The Department of Electrical Engineering at Tri-State College is all one entity

\item Do not include extra punctuation or spaces in the labels, unless they occur within a named entity. If OCR errors produce ``First (Baptist Church'', the extra parenthesis should be included since it's inside a named entity. But if you have ``(First Baptist Church'', do not include the parenthesis. 

\item On a similar note, given something like ``Albert Sealy's cat'', only label ``Albert Sealy''. Do not include the apostrophe and the ``s''.

\item Where relevant, don't label the ``the'' (ie. should be ``State Department'' not ``the State Department''). 

\item For newspapers and other publications/organizations use Google to see if “The” is part of their official name. For instance, ``The New York Times'' is the actual full name of the newspaper, so label the ``The''. However, ``The Bell Syndicate'' doesn’t actually have ``The'' as part of their official name, so just label ``Bell Syndicate''. 

\end{itemize}

\subsection{Locations}

\begin{itemize}

\item If a location title is used as an adjective, it is miscellaneous. e.g. ``US senator'' should have ``US'' labelled as MISC, not LOC.

\item If there is an organization/location ambiguity, we default to labeling it as a location. If the `location' does an action, it is labeled as an organization. e.g. White House = LOC unless ``the White House'' does something. e.g. I’m going to the Natural History Museum and the Marriot hotel are both locations.

\item EXCEPTION: All administrative units (countries, states) are locations and never organizations, even when doing an action.

\item If a location is part of the name of an organization, we label the whole thing as an organization (e.g. ``the church of christ in America'' or the ``Ohio agriculture department'').

\item We’re defining locations as 4D: distinctive time periods that are capitalized for places are included in locations e.g. Victorian England, Ancient Greece, Nazi Germany, and Red/Communist China are all labeled as single locations.

\item Cardinal direction plus location: if defined in popular speech as a specific location, label the direction also e.g. Western Europe, Central London.

\item If the secondary word is plural, it is not a specific location and is not included in the label (e.g. Marriott hotels).

\item When established locations are used to reference a location in relation to them, the established locations are labeled separately. E.g. “between Smith and Adams street” are two separate locations.

\item This also applies to railroad lines e.g. “the Chicago-New York line” should have Chicago and New York labeled separately.

\item Named objects in space (e.g. Sun, Moon, Jupiter) are miscellaneous unless someone/something is going to it or if there is a reference to a place on the object, then they are locations.
\end{itemize}

\subsection{People}
\begin{itemize}

\item For person entities, do not label Mr, Mrs, and other prefixes (such as Dr, Rev, etc.) and don’t include suffixes (e.g. ``Jr.'', ``III'').

\item Similarly, do not label descriptors like ``Deputy Sheriff'' that often come before people's names.

\item Also, positions (without a name) are not named entities e.g.: The Minister of Foreign Affairs, Maharajah, the Queen should not be labelled.

\item Include any nicknames when they are put in the middle of names (e.g. something like ``Dwayne `The Rock' Johnson'' should be all labeled person).

\item God is a person, but pronouns like ``He'' and ``Him'' are not.

\item Animal names are miscellaneous, not person.

\end{itemize}

\subsection{Organizations}

\begin{itemize}

\item If an organization is used as an adjective, it is miscellaneous.

\item Groups of people that go by a name (e.g.: Republicans, Cavaliers) can be orgs, but only if we’re referring to the whole group (e.g. ``the republicans'' referring to the whole party, not a group of five republicans).

\item Liberals and conservatives are not organizations (at least in the US), since the definitions often change and are not really groups.

\item Many politician names will be followed by (D/R-State). Label the D/R as miscellaneous, since the group is being used as an adjective to describe the politician. Label the state as location.

\item If there is an organization/location ambiguity, we default to labeling it as a location. If the ``location'' does an action, it is labeled as an organization e.g. `the ``White House'' reported,’ is an organization, but `I’m going to the Natural History Museum' is a location.

\item All administrative units (countries, states) are locations and never organizations, even when doing an action.

\item Brands of products are organizations if they are the name of the company/producer (e.g. Apple computers) or unless they are now ubiquitous (e.g. ziploc bag, polo shirt). If the brand is not the name of the company, it’s miscellaneous. E.g.: Toyota is an org, highlander is misc, x-ray is neither, 45-caliper gun is neither.

\item Ambiguous organizations are still labeled when they are non-ambiguous in context (e.g. an article talking about Cleveland saying the ``rotary club'').

\item This also applies to slightly ambiguous government entities like the “army”, “navy”, or “state department” - if it’s clear it’s the US entity in the article, label it as such.

\end{itemize}

\subsection{Miscellaneous}

\begin{itemize}

\item Adjectives derived from named entities are miscellaneous named entities e.g. nationalities used as adjectives (e.g. U.S., French, London Newspapers).

\item EXCEPTION: People used as “adjectives” (e.g. “the Kennedy household”) are to be labeled as PER, not MISC.

\item EXCEPTION: When a person’s name has become part of a famous location (e.g. “Eiffel Tower”, “Chandler Building”) that has its own Wikipedia page (or equivalent) the entire location is considered an entity and labeled appropriately.

\item When an entity is used as a possessive, that is with an apostrophe or “of” (e.g. “Wisconsin’s cows”, “people of France”), the entity should be labeled with its original label, not MISC.

\item Congressional, senatorial, constitutional are all not considered adjectives from named entities.

\item Political ideologies are miscellaneous (e.g. communist, socialist, conservative, authoritarian etc.).

\item Include prefixes and suffixes to these in misc labels e.g. anti-Japanese, pro-communist.

\item Names of groups of people/religions are miscellaneous (only if they’re capitalized or `should' be capitalized, e.g. ``communists'' should be misc, but ``visitors'' should not be).

\item Titles/positions are not miscellaneous named entities e.g.: The Minister of Foreign Affairs, Maharajah, the Queen are not misc.

\item Officially named initiatives/programs are misc e.g. Manhattan Project, U.S. Census.

\item Names of capitalized documents or forms are miscellaneous e.g. Individual Census Reports, the Constitution.

\item Distinct political acts (e.g. ``Agricultural Act of 1970'') are also misc.

\item Capitalized/specific names of objects are misc e.g. USS Canopus.

\item Names of animals are miscellaneous e.g. Laika (the space dog).

\item Events are miscellaneous.

\item Events must be famous or distinct (e.g. “Pearl Harbor”, “1969 World Series”).

\item Less famous events must refer to something occurring within a specific timeframe (e.g. there will be a ``city council meeting'' is not a named entity). The event should be unambiguous (e.g. the “city council meeting on 11/5” is not a named entity because it doesn’t specify which city council but the ``Boston City Council meeting on 11/5'' is).

\item Christmas and other major holidays are miscellaneous.

\item Brands of products are organizations if they are the name of the company/producer (e.g. Apple computers) or unless they are now ubiquitous (e.g. ziploc bag, polo shirt). If the brand is not the name of the company, it’s miscellaneous. e.g. Toyota is an org, highlander is misc but not org, x-ray is neither, 45-caliper gun is neither.

\item Named objects in space (e.g. Sun, Moon, Jupiter) are miscellaneous unless someone/something is going to it or if there is a reference to a place on the object, then they are locations.
\end{itemize}

\subsection{NER Results}
The following shows the breakdown of shares of different types of entities (in terms of overall tokens) from applying the NER model to newswire articles from the 20th century. The figure shows broad stability in entity mentions across time, with the exception of World War II, when location and miscellaneous entities (\textit{e.g.}, such as named aircraft) spike. 

\begin{figure}[h]
    \centering
    \includegraphics[width=.95\linewidth]{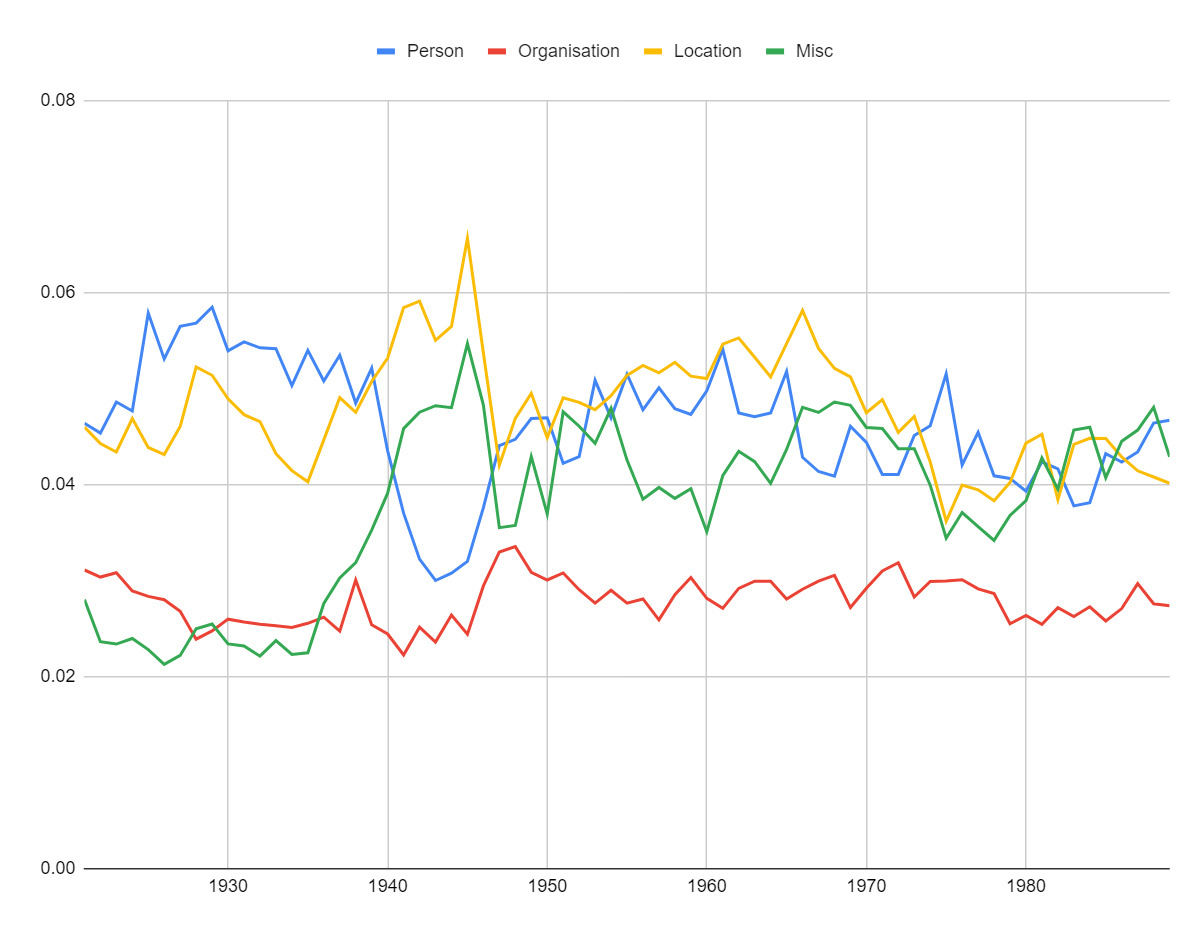}
    \caption{Shares of entity types in newswire articles.} 
    \label{fig:ner}
    \vspace{-4mm}
  \end{figure}

\clearpage

\section{Explanation and Examples for Evaluation of \ndjv}

\subsection{Explanation}

The \ndjv package was used to obtain a total of 5 historical articles from our corpus per modern article of which there were 70. These 70 articles were pulled from the websites of popular news outlets such as the Associated Press, Fox News, USA Today, and the Washington Post. This was done using the \ndjv python package. Each modern article was then placed in a pair with its 5 associated historical articles. A skilled annotator was then tasked with manually classifying the articles in a modern-historical pair as being on or off the same topic.

Being on or about the same topic was defined in the following way: two articles are about the same topic if absent named entities (e.g., proper nouns), the relationship between the most important remaining concepts are essentially the same for both articles. Any two articles that fail to meet these criteria according to the skilled annotator were deemed off or not about the same topic.

Two articles being about the same topic can also be thought about in relation to two articles being on the same story and or on the same events. Topics, stories, and events are distinguished along two dimensions: time and named entities. For two articles to be about the same event, what they each describe must have occurred during the same time period (no more than a day) and have the same or closely related named entities. To be about the same story, two articles must have the same or closely related named entities, but the actions off named-entities or the relationships between named-entities may have occurred over long stretches of time. In both events and stories, named-entities serve the same function as a fictional story’s cast of characters. Named-entities and when their actions occur are irrelevant for determining two articles’ shared topic. Events, stories, and topics are also nested concepts. Every event is part of a story and every story is part of a topic. An article may also belong to more than one event, story, and/or topic.

An illustrative example might be “Watergate”. The “Watergate Break-In” was an event that occurred on Saturday, June 17, 1972 at the Watergate Hotel in Washington, D.C. and was foiled by a security guard named Frank Wills. This is an event because it occurred on a discrete date in time with a clear set of named-entities, like the Watergate Hotel, Washington, D.C., and Frank Wills. If another article was about perpetrator Virgilio Gonzalez’s arrest at the Watergate Hotel that night, that article would also be about the “Watergate Break-In” because the arrest occurred at the same date in time. Virgilio Gonzalez is related to the other named entities. The Watergate Hotel in D.C. was the location of his arrest, and it was Frank Willis’ tip to the police that led to his arrest. The “Watergate Scandal” would be a story as it unfolded over years and included named entities that were related to one another but did not all participate in the same events. For example, Richard Nixon did not personally break into the Watergate Hotel, but it is believed that he tried to cover up his connections to the break-in’s perpetrators, like Virgilio Gonzalez. The cover-up also occurred sometime after the break-in. An article about Richard Nixon’s cover-up activities would be about the same story as an article about Virgilio Gonzalez’s arrest. “Political Scandals” would be an example of a topic that the “Watergate Scandal” falls under. An article about the “Watergate Scandal” and an article about President Donald Trump pressuring election officials for more votes would both be political scandals. The named entities and dates in time are different, but the most important remaining concepts, like accusations of election interference against presidents, are essentially the same.

\subsection{Examples of Historical-Modern Article Pair Evaluation}

The following articles make five of the labeled article pairs, each containing the same modern article and one of the five historical articles retrieved using the \ndjv package. Three historical articles were classified as being on the same topic and two were classified as being on different topics. In the same way that you can provide names for events, stories, or topics, like “Political Scandals”, the skilled annotator was tasked with doing so for the evaluation of the modern-historical article pairs and those names are included here. In order to abide by copyright restrictions, the full modern article is not reproduced here, just a truncated version. A link to the full article is provided however. Due to OCR errors, the historical article text may appear less coherent than the modern article but still readable nonetheless.

\subsubsection{Modern Article}

\paragraph*{\\Modern Article URL\\\\}

\url{https://www.usatoday.com/story/money/food/2024/03/13/ben-jerrys-free-cone-day-2024/72944410007/}

\paragraph*{Modern Article Headline\\\\}

Ben \& Jerry's annual Free Cone Day returns in 2024: Here's when it is and what to know

\paragraph*{Modern Article Body\\\\}

Ben \& Jerry's is bringing back its annual Free Cone Day celebration this spring and is asking fans to help them beat a lofty goal.

The company wants this year's Free Cone Day to be the "biggest and best yet with 1 million scoops served," it announced Wednesday.

This year's celebration will take place on Tuesday, April 16, the company said in a news release. Free Cone Day made its return last year after a four-year hiatus.

\subsubsection{Historical Article 1}

\paragraph*{\\Historical Article ID\\\\}

10\_304477967-ottumwa-daily-courier-Jan-04-1943-p-1.jpg

\paragraph*{Historical Article Headline\\\\}

Additional Cut In Ice Cream Output

\paragraph*{Historical Article Body\\\\}

Washington, D. C.-(P)-The war production board today limited January production on ice cream to 50 per cent of the amount each manufacturer made last Ocotber.

- This represented a reduction of one sixth from December, when each manufacturer was permitted to make 60 per cent of his October amount.

The order also applies to frozen custard, milk sherbet, other frozen desserts and ice cream mix.

W.P.B. said the order “wag ise sued at the request of United States department of agriculture . . » to further relieve the butter shortage.”

\paragraph*{On Topic with Modern\\\\}

\text{True}

\paragraph*{Topic Name / Notes\\\\}

Ice Cream

\subsubsection{Historical Article 2}

\paragraph*{\\Historical Article ID\\\\}

1\_106022053-titusville-herald-Apr-11-1939-p-1.jpg

\paragraph*{Historical Article Headline\\\\}

52,299 Jam White House Lawn ‘For Annual Easter Ego Rolling

\paragraph*{Historical Article Body\\\\}

OY 2£2RE NAAGCIAER Pear

WASHINGTON, April 10.—Even Alice {in Wonderland couldn't have dreamed anything like the Easter rolling which took 52,259 children and adults—by the gate keepers’ count-—to the private grounds of the White House today.

Through the gates streamed a tenyear-old boy dressed like a white rabbit with flapping pink ears... a live white rabbit scampering on a leash -. . . \@\& huge chocolate bunny perched beside a grinning infant in a baby carriage.

Bands played. A magician did tricks. Crowds gathered about a Punch and Judy show. Everywhere,- there were children and eggs—eggs crunched tnderfoot, smeared on young faces, salling between stately old White House trees and rolling down slopes.

It was like a glorified country pienic. Thousands spread their lunches on the ground. Then tt wasn't, for there on the White Mouse poriica wag Lhe Pres

ident of the United States, waving and wishing, he said, that he could “be down there with you.”

And four times during the pleasant sunny day the country's gracious First Lady appeared. Three times she made trips through the grounds, smiling, | waving and calling “how-do-you-do” to those along her path. | As she walked she saw children crawl; about on all fours, rolling themselves down slopes, sitting spraddie-legged on the ground to juggle eggs and swingIng from iow limbs of trees.

It was the seventh time, the President pointed out, that he and Mrs. Roosevelt had entertained the children at the White House lawn. The roll has been an annual event at the White House, except for the World war years, since 1878.

“It's a wonderful day,” he declared hefore he disappeared into the house, “and I hope you u enjoy. yourselves very, much,?

\paragraph*{On Topic with Modern\\\\}

\text{False}

\paragraph*{Topic Name / Notes\\\\}

one is about ice cream and the other is about Easter

\subsubsection{Historical Article 3}

\paragraph*{\\Historical Article ID\\\\}

1\_10883667-daily-globe-Mar-01-1949-p-1.jpg

\paragraph*{Historical Article Headline\\\\}

Butter vs. Oleo Battle, Annual Congress Feature, Starts Today

\paragraph*{Historical Article Body\\\\}

Washinglon — ¢Yfj—The butter vs. oleo battle, an annual feature of congress, got underway today,

Rep, Poage (D-Tex) fired the first shot with an attack on “the butter lobby.” He said it is more interested in tryilng "to hill yelJow margarine” than it is in protecting consumers from fraud.

Poage was the first witness at house agriculture committee bhearmgs on 29 bills regarding margarine, Some of them would repeal the taxes on margarine, but prohibit the manufacture und sale of yellow margarine,

Poage sald this latter proposal is an effort en the part of the butter industry “to appear to yield to outraged public sentiment against inexcusable favoritism In favor of one wholesome food

-aguinst anolher without actually lgiving up any af the special pri vilege butler” bas so long enjoy: ed."

|POAGE'S BILL

Poage has a bill which would lift, federal taxes on margarine’ and permit manufacturers 10 col: or it yellow. But it would require public eating places to notify cus: tomers \$f they serve margarine.

The issue as whether there should be a federal tax on the

| butter substitute. The government ;Now taxes all of it, with on ex. tra toll if at is ecalored

President Truman for repeal of the margarine ti ;@s, and so did the Democri platform on which he ran list fall, The Republican platform did ; not mention the subject.

Tf the controversy follows its new almost traditional form — butter partisans will suggest that margarine could be colored anything from purple or green to a bright, cherry red=-anything but yellow,

The margarine folks will say butter has no exclusive cham to yellow, and in fact the yellow color has to be added ta some bulter.

Southern Democrats from rural districts and northern Democrats sand even Republicans — from big city districts will ally for the moment in support of marparine,

There are two types of Dllls before ihe commiltee,

The most numerous would abolish all federal taxes on margnrine. Others would but out st least the added nx on colored margare.

At the other end of the line are bis by Reps. August 11. An ;aresen (R-Minn), und Granger '(D-Uiah) to prohibit the manufacture and sule of yellow marigaring, STARTED 69 YEARS AGO

The whole thing started 63 years ago, when the first federal anti-margarine Jaw was pussed. That law and others which followed have been under attack almost every year since.

The oleo forees made their grealest progress last yeat, when a bill fo repeal the margarine taxes passed the house handily. It never got to a vote in the sen | ate.

A federal tax of 10 cents a pound is now paid on all colored margarine sold at retail. The retail tax on the uncolored product is one-quarter cent a pound.

Margarine manufacturers pay a federal tax of \$600 a year; wholesalers pay a tax of \$480 4 year Sf they handle colored margarine, and \$200 a yenr for uncolored, and retailers are taxed \$48 n year fo handle colored mare garine and \$6 10 handle uncol

\paragraph*{On Topic with Modern\\\\}

\text{True}

\paragraph*{Topic Name / Notes\\\\}

food

\subsubsection{Historical Article 4}

\paragraph*{\\Historical Article ID\\\\}

26\_90193365-morning-herald-Apr-04-1946-p-1.jpg

\paragraph*{Historical Article Headline\\\\}

No More KP In The Air Forces Of Army

\paragraph*{Historical Article Body\\\\}

Washington, April 3 (P) ~ There will be no more KP (Kitchen Police) duty in the Army Air Forces under a new program announced today.

Soidiers will stilt peel spuds and wash dishes, But those who do will be permanently assigned to the task and will be called “mess attendants.” The anMouncement adds that they “will be afforded an opportunity to make an Army career of food service,”

The old system of assigning all men on the roster to KP in turn is being abolished.

The AAF announcement said that “many (ocal exigencies and personnel problems” prevent setting a definite date for the establishment of Utapia.

\paragraph*{On Topic with Modern\\\\}

\text{True}

\paragraph*{Topic Name / Notes\\\\}

Food

\subsubsection{Historical Article 5}

\paragraph*{\\Historical Article ID\\\\}

5\_233435376-circleville-herald-Dec-30-1976-p-1.jpg

\paragraph*{Historical Article Headline\\\\}

International Falls No Sunny Spa

\paragraph*{Historical Article Body\\\\}

INTERNATIONAL FALLS, Minn. (AP) — At the close of a year, a time for reflection, hardly a place is better suited than this for that worthy exercise.

There is something to be learned from this place, something other than what everybody already knows from the nightly weather report: that it is the coldest place in the 48 contiguous states.

When winter’s fangs bite into this little spot on the Canadian border — in the first half of this month the thermometer managed to hang above zero for only four brief hours — living becomes an adventure and humility a daily lesson. Nature’s elemental severity invites long thoughts about

man’s standing in the Great Scheme.

“IT think we worry more about the simple necessities of survival than most people do,” said Frank Bohman, a philosophic aviator who has lived here all his 52 years.

‘Having enough food in the house, enough fuel, a backup heating system, these are real concerns. I would imagine that in gentler climates people take survival for granted.”’

For the record, when the earth tilts toward winter, winds borne on the jet stream sweep from the North Pole down the interior flank of the Canadian Rockies and pivot eastward right at this point, so that the average yearly temperature here is 37.5 degrees and the annual snowfall 50 inches. Readings in the minus 30s and 40s are commonplace during the winter.

The cold grips so fiercely, in fact, that it all but refuses to let go. The ground freezes five feet down, untillable until June.

The town is on the granite shore of Rainy Lake, one of creation’s masterpieces, a 340-square-mile work of art done in a freeform of coves and bays and flecked with 1,600 tiny granite islands timbered with pine.

Thus in the summertime the area is awash with tourists, regulars who return to their summer places on the islands, weekenders seeking walleyed pike and clean air, visitors with cameras, water skis and time to make the two-hour drive up from Duluth.

When the summer crop of frolickers is harvested, however, only a bold band of the hearty remain to face the long dark winter.

That yearly experience has given them a palpable sense of neighborliness, a closeness such as a shared secret brings. When the ice breaks up each May they have earned a communal handshake that says nice going everybody, we did it again, we didn’t quit.

Those brave souls number 9,109 in. International Falls and the nearby communities of South International Falls and Ranier. About the same number live across the Rainy River Bridge at Fort Frances, Canada.

The spirit of hands-across-the-sea, or in this case, the river, comes naturally; nature's legacy knows no international boundary. Indeed, one longtime Chamber of Commerce president in International Falls, Gordy McBride, was a Canadian citizen.

\paragraph*{On Topic with Modern\\\\}

\text{False}

\paragraph*{Topic Name / Notes\\\\}

one about ice cream and the other about weather

\end{document}